\documentclass{article}
\usepackage{graphicx, xcolor} 
\usepackage{amsmath,amsfonts,amssymb,amsthm,enumerate,mathtools,bbm}
\usepackage{tikz}
\usetikzlibrary{arrows.meta,decorations.pathreplacing}
\usepackage{capt-of} 
\usepackage[main,final,nonatbib]{neurips_2026}
\usepackage{hyperref}
\usepackage[normalem]{ulem}
\usepackage[colorinlistoftodos]{todonotes}

\usepackage{soul}
\usepackage{comment}
\usepackage{enumitem}
\let\epsilon\varepsilon
\usepackage[style=alphabetic]{biblatex}
\newtheorem{theorem}{Theorem}[section]
\newtheorem{lemma}[theorem]{Lemma}
\newtheorem{lemma*}[theorem]{Lemma*}
\newtheorem{theorem*}[theorem]{Theorem*}

\newtheorem{corollary}[theorem]{Corollary}

\newtheorem{remark}[theorem]{Remark}

\theoremstyle{definition}
\newtheorem{definition}[theorem]{Definition}

\DeclarePairedDelimiter{\floor}{\lfloor}{\rfloor}

\DeclareMathOperator*{\argmax}{arg\,max}

\newcommand{\R}{\mathbb{R}}
\newcommand{\N}{\mathbb{N}}
\newcommand{\Z}{\mathbb{Z}}
\newcommand{\Ncal}{\mathcal{N}}
\newcommand{\x}{\mathbf{x}}
\renewcommand{\a}{\mathbf{a}}
\renewcommand{\o}{\mathbf{o}}
\renewcommand{\b}{\mathbf{b}}

\newcommand{\q}{\mathbf{q}}
\renewcommand{\k}{\mathbf{k}}
\renewcommand{\v}{\mathbf{v}}
\newcommand{\y}{\mathbf{y}}
\newcommand{\z}{\mathbf{z}}

\newcommand{\W}{\mathbf{W}}
\newcommand{\M}{\mathbf{M}}
\newcommand{\I}{\mathbf{I}}
\newcommand{\0}{\mathbf{0}}
\newcommand{\layernorm}{\text{layer\_norm}}
\newcommand{\multilayernorm}{\text{multi\_layer\_norm}}
\newcommand{\identity}{\text{identity}}
\newcommand{\hardmax}{\text{hardmax}}
\newcommand{\mask}{\text{mask}}
\newcommand{\CoT}{\mbox{\rm CoT}}

\newcommand{\WE}{\text{WE}}
\newcommand{\PE}{\text{PE}}
\newcommand{\1}{\mathbbm{1}}
\newcommand{\Att}{\text{Att}}
\newcommand{\ReLU}{\text{ReLU}}

\newcommand{\TIME}{\mbox{\rm TIME}}

\newcommand{\polylog}{\text{polylog}}

\title{A Tight Hierarchy for Chain of Thought}
\author{
  Thanh Le \\
  Department of Mathematics \\
  University of Nebraska--Lincoln \\
  \texttt{tle51@huskers.unl.edu}
  \And
  A. Pavan \\
  Department of Computer Science \\
  Iowa State University \\
  \texttt{pavan@iastate.edu}
  \And
  N. V. Vinodchandran \\
  School of Computing \\
  University of Nebraska--Lincoln \\
  \texttt{vinod@unl.edu}
}
\date{April 2026}

\makeatletter
\renewcommand{\@noticestring}{%
  A version of this paper will appear at NeurIPS 2026.%
}
\makeatother

\begin{document}

\maketitle


\begin{abstract}
Chain-of-thought (CoT) reasoning has emerged as a powerful paradigm for extending the computational capabilities of transformer-based models. A growing body of recent work has begun to uncover the fine-grained computational power of transformers augmented with CoT, as well as their connections to classical complexity theory.

In this work, we advance this line of research by establishing the first hierarchy theorem for \CoT-based computation. Our main result shows that for any reasonable resource bound $r(n) \geq n$  and any $\varepsilon > 0$,
\[
\mathrm{CoT}(r(n)) \subsetneq \mathrm{CoT}(r(n)^{1+\varepsilon}),
\]
where $\mathrm{CoT}(r(n))$  denotes the class of decision problems solvable by transformers using $O(r(n))$ CoT reasoning steps on inputs of length $n$. Our result shows that increasing the number of CoT reasoning steps provably yields strictly greater computational power, establishing CoT steps as a fundamental resource in transformer-based computation.

Unlike classical hierarchy theorems, which are typically proved via diagonalization, our approach leverages known coarse-grained simulations between CoT-augmented transformers and Turing machines in both directions. Building on these connections, we adapt the classical padding technique from complexity theory to this setting to obtain a tight hierarchy. A key technical contribution of our work is the implementation of padding within the transformer-based CoT model.
\end{abstract}



\section{Introduction}

A central objective in computing foundations is to understand how computational power scales with access to additional resources. The classical hierarchy theorems from computational complexity theory establish that increased time or space yields strictly greater computational capability. The two most fundamental results in this regard are the Time Hierarchy Theorem and the Space Hierarchy Theorem. Intuitively, the Time Hierarchy Theorem states that if $t(n) \geq n$ is a sufficiently nice complexity-bounding function (i.e., time-constructible function), then there are problems solvable in time slightly more than $t(n)$ that cannot be solved in time $t(n)$. In particular, the most well-known version of this principle states that 
$\TIME(t_1(n)) \subsetneq \TIME(t_2(n))$
if $t_1(n)\log t_1(n) = o(t_2(n))$~\cite{hartmanis1965computational}.
Here $\TIME(t(n))$ is the class of languages that are decidable in time $O(t(n))$ on inputs of length $n$ using a multi-tape Turing machine. Hierarchy theorems 
formalize a basic structural principle: bounded resources induce strict stratifications of computational power. Even though hierarchy theorems are intuitive, it is far from trivial to establish such quantitative results. Without such results, we cannot rule out the possibility of universal thresholds on resources that are required to solve any decidable problem.  

Modern AI architectures, particularly transformer architectures, have introduced new models of computation that are both practically powerful and theoretically intriguing. Transformers, built on mechanisms such as self-attention and layered representations, have achieved remarkable success across domains including natural language processing, vision, and scientific computing. The success of transformer architectures led to 
a new line of research to understand their computational power. Recent research formulated this question by modeling transformers as a computation device, thus enabling rigorous study of the power and limitations of this new model. 

A growing body of work  has identified fundamental limitations of standard transformer architectures \cite{hahn2020theoretical,chiang2023tighter,liu2023transformers_shortcuts,hao2022formal,merrill2022saturated,merrill2023expressive,merrill2023parallelism}. In particular, under realistic assumptions, constant-depth transformers are less powerful than finite state machines, arguably the simplest computational model. A recent survey \cite{Strobl_2024} gives an excellent account of the expressive power of transformers as a computational device. The intuitive reason is that constant-depth transformers operate in a largely feedforward manner and lack the ability to perform the iterative computation required for multi-step reasoning. As a result, even tasks that are conceptually simple but require sequential composition of intermediate steps fall outside their effective computational capabilities~\cite{PengNarayananPapadimitriou2024,DziriLSLJLWWB0H23,ZhangPMLS24}. 

Modern AI systems are augmented with interactive reasoning steps where the output of a step can be used to further the computation. This is known as chain-of-thought (CoT) reasoning. Chain of thought (CoT) refers to a mode of computation in which a model produces explicit intermediate steps before arriving at a final answer. Instead of mapping an input directly to an output in one pass, the model generates a sequence of reasoning states that progressively transform the input into the solution. 

From a computational perspective, this modification significantly alters the computational landscape. A growing body of work shows that the number of such intermediate reasoning steps plays a crucial role in determining expressivity. Thus, chain of thought emerges not merely as a heuristic for improving empirical performance, but as a fundamental computational resource in transformer-based models. In particular, Perez et al.\ show that this model is Turing complete in the sense that anything that is computable using a Turing machine is computable by transformers with enough \CoT\ steps~\cite{perez2021attention}. This connection is made more explicit by Merrill and Sabharwal~\cite{merrill2023expressive}. In particular, they showed that any language recognized by $t(n)$-time bounded Turing machine can also be recognized via a transformer with $O(t(n))$ \CoT\ steps and any transformer with $O(t(n))$  \CoT\ steps can be simulated by a Turing machine in time $O(t^2(n)\polylog(t(n)))$. 

In this work, we propose to view the number of \CoT\ steps as a resource (such as time and space in TM model) and ask the fundamental question: 
\begin{center}
\em Is there a strict hierarchy for \CoT\ reasoning?
\end{center}
Let $\CoT(r(n))$ be the class of problems that can be decided by transformers with {$O(r(n))$} chain-of-thought steps. 
Our main result is that for any reasonable function $r(n)$, and for any $\varepsilon > 0$,
\[
{\CoT}(r(n)) \subsetneq {\CoT}(r(n)^{1+\varepsilon}).
\]
Our result shows that chain-of-thought reasoning admits a fine-grained hierarchy: a small increase in the number of reasoning steps strictly increases computational power. This establishes \CoT\  as a meaningful computational resource similar to time and space.

A classical technique underlying hierarchy theorems is diagonalization. In the setting of Turing machines, diagonalization constructs a language that disagrees with every Turing machine running within a given resource bound, by simulating each such machine on carefully chosen inputs and flipping its output. This approach relies on two key properties: (i) machines can be effectively enumerated, and (ii) there exists a universal machine that can take the description of another machine as input and simulate its execution with controlled overhead. These properties make it possible in the Turing machine model to carry out self-referential constructions in a clean and uniform way.

For transformers with chain of thought, the situation appears to be different. While one can, in principle, encode a transformer as a finite object (e.g., via its architecture and parameters), there is no conceptually clean notion of what it means for a transformer to take another transformer as input and simulate its computation. Transformer computation is specified by a fixed architecture, layers, attention heads, and learned weights—rather than a discrete, easily interpretable transition function. As a result, even defining a “universal transformer” that can parse such an encoding and simulate another transformer’s behavior in a step-by-step manner is conceptually unclear. This lack of a natural encoding-and-simulation approach makes it difficult to carry out the kind of self-referential constructions that diagonalization relies on.


\subsection{Our Approach}

Our approach builds on known simulations relating chain-of-thought computation and classical time-bounded computation. In particular, \cite{merrill2023expressive} establishes the following: for any $t(n) \geq n$ (1) $\CoT(t(n)) \subseteq \TIME(t^3(n))$\footnote{They prove a slightly stronger containment, but the stated bound suffices for our purposes.} and (2) $\TIME(t(n)) \subseteq \CoT(t(n))$. By combining these two simulations and using the classical time hierarchy theorem, we can get a coarse hierarchy theorem for \CoT\ as follows. 
\[\CoT(t(n)) \subseteq \TIME(t^3(n)) \subseteq \CoT(t^3(n)) \subseteq \TIME(t^{9}(n)) \subseteq \CoT(t^9(n)) \]
This chain of containments implies that $\CoT(t(n)) \subsetneq \CoT(t^9(n))$: if not, $\TIME(t^3(n)) = \TIME(t^9(n))$, contradicting the Time Hierarchy Theorem. However, this approach only gives us a very coarse hierarchy theorem. In particular, it will not separate $\CoT(n)$ from $\CoT(n^2)$, let alone separate $\CoT(n)$ from $\CoT(n^{1.01})$. A way to obtain such separation is via a very efficient simulation of transformers by a Turing machine, such as $\CoT(t(n)) \subseteq \TIME(t^{1+o(1)}(n))$.  However, each decoding step of a transformer attends to all preceding tokens. This incurs a quadratic cost to simulate via Turing machines.

To obtain sharper results, we use the {\em padding technique} from complexity theory. 
The padding technique enables the transformation of coarse separations into finer-grained separations.
By adapting padding to the transformer model, we are able to derive a tighter separation:
$\CoT(r(n)) \subsetneq \CoT(r(n)^{1+\varepsilon})$. 
The main technical challenge lies in adapting the padding technique to the transformer setting, where inputs and intermediate representations must be encoded within the constraints of the model. This is our primary technical contribution. 
\subsection{Related Work}
In addition to the works discussed in the introduction on the expressive power of transformers with chain-of-thought reasoning, we briefly review classical hierarchy theorems from complexity theory. The time hierarchy theorem was established in the seminal work of Hartmanis and Stearns~\cite{hartmanis1965computational}. This was followed by the \emph{space hierarchy theorem} in~\cite{StearnsHartmanisLewis1965}. 
These results were subsequently extended to nondeterministic computation. The \emph{nondeterministic time hierarchy theorem}~\cite{Cook1973,SeiferasFischerMeyer1978} shows that
$
\mathrm{NTIME}(t(n)) \subsetneq \mathrm{NTIME}(t'(n)),
$
for suitably larger $t'(n)$, while a corresponding \emph{nondeterministic space hierarchy theorem} was established by Ibarra~\cite{Ibarra1967}. Establishing hierarchy theorems is still an active area of research in complexity theory, for example for randomized computations there has been only partial progress~\cite{Barak02,FortnowS04,FortnowST05}.

Recent works have studied transformers as formal computational models, relating their expressive power to automata, circuit complexity, and classical complexity classes. The work of~\cite{perez2021attention} showed that transformers with hard attention and arbitrary precision are Turing complete. Subsequent work focused on understanding the power of transformer variants under bounded precision and architectural restrictions. The work of~\cite{hahn2020theoretical} showed that constant-depth transformers cannot recognize several basic formal languages, including PARITY and Dyck languages. The work of~\cite{hao2022formal} proved hard-attention transformers can be simulated by constant depth circuits (belonging to the complexity class $\mathrm{AC}^0$). In~\cite{merrill2022saturated}, this result has been extended to saturated attention transformers by providing an upper bound of $\mathrm{TC}^0$.
 Other works further explored limitations arising from bounded depth, restricted attention, and finite precision~\cite{chiang2023tighter,liu2023transformers_shortcuts,merrill2023parallelism,merrill2023expressive,PengNarayananPapadimitriou2024}. More recently, the role of intermediate decoding and chain-of-thought reasoning has received attention~\cite{merrill2023expressive,ChenPW25}. We refer the reader to~\cite{Strobl_2024} for a comprehensive survey of this rapidly growing area.

\section{Preliminaries}
We consider \textit{decoder-only transformers} with chain of thought that can generate intermediate tokens between reading the input and generating an answer. Our model mostly follows definitions as in \cite{merrill2023expressive}. 
In the main body of the paper, we describe the main points about the model. We start here with notation that we use throughout the paper.  

 Let $[n] := \{1, 2, \dots, n\}$. A function $t: \mathbb{N} \rightarrow \mathbb{N}$ is time-constructible if there is a Turing machine $M$ that on every input length of $n$ runs for exactly $t(n)$ steps and stops. All standard time functions, such as polynomials and exponentials, have this property. We (most of the time) use lowercase letters ($x, y,\dots$) for tokens in an alphabet $\Sigma$; uppercase letters ($X, Y, \dots$) for strings in $\Sigma^*$. We use lowercase bold letters ($\mathbf{x}, \mathbf{y}, \dots$) for vectors; uppercase bold letters ($\mathbf{X}, \mathbf{Y}, \dots$) for matrices. We work with vectors which can be either a row matrix or a column matrix. Sometimes, we do not specify row or column for conciseness -- whether a vector is a row or a column should be clear from context. We write ``$\identity$" to be the identity map, i.e. for any vector $\x$, $\identity(\x) = \x$. In terms of matrix, we write $\I_{n \times n}$ for the $n \times n$ identity matrix. We write $\0_{m \times n}$ for the $m \times n$ zero matrix. We write ``$\_$" to mean ``some number in $\R$," and likewise we write $(\_)^d$ to mean ``some vector in $\R^d$". 

 We use indicator-notation: if $\phi$ is a proposition, we write 
    \begin{equation*}
        \1(\phi) = \begin{cases}
            1 &\text{if $\phi$ is true} \\
            0 & \text{otherwise.}
        \end{cases}
    \end{equation*}






\begin{definition}
Let $\Sigma$ be a finite alphabet containing a designated accept symbol $1 \in \Sigma$. 
For a function $r:\mathbb{N}\to\mathbb{N}$, a transformer $T : \Sigma^* \to \Sigma$ \textit{decides }a language 
$L \subseteq \Sigma^*$ in $r(n)$ chain-of-thought steps if the following holds.

On input $X \in \Sigma^n$, define a sequence of strings $\{f^k(X)\}_{k\ge 0}$ by
$f^0(X) = X,
f^{k+1}(X) = f^k(X)\cdot T(f^k(X)),
$
where $\cdot$ denotes concatenation. Each $T(f^k(X))$ is called a \textit{decoding step}. After $r(n)$ decoding steps, the output of $T$ is the final token of $f^{r(n)}(X)$. We say that $T$ accepts $X$ if this output equals $1$. The language decided by $T$ is
\[
L = \{X \in \Sigma^* : \text{the final token of } f^{r(|X|)}(X) \text{ is } 1 \}.
\]

We define $\CoT(r(n))$ to be the class of all languages $L \subseteq \Sigma^*$ that are decided by some transformer $T$ using at most $O(r(n))$ decoding steps.
\end{definition}


Let $\TIME(t(n))$ denote the class of decision problems decidable in $O(t(n))$ time steps using a multi-tape Turing machine. Let $\widetilde{\TIME}(t(n))$ denote $\TIME(t(n)\polylog(t(n)))$. \cite{merrill2023expressive} shows the following relations between \TIME\ and \CoT.  

\begin{theorem}[\cite{merrill2023expressive}] \label{thm:hierrachy TIME vs CoT in MS23}
    With the definition above, we have that 
    \begin{equation*}
        \TIME(r(n)) \subseteq \CoT(r(n)) \subseteq \widetilde{\TIME}(r^2(n) + n^2).
    \end{equation*}
\end{theorem}
In particular, \cite{merrill2023expressive} assumes the following: $log$-precision (i.e. for input of length $n$ and $r(n)$ intermediate decoding steps, precision is $O(\log(n + r(n)))$); saturated attention; and multi-layernorm. We make the same assumptions for our constructions, together with a few reasonable additional assumptions. More details on model assumptions are explained in Appendix \ref{sec:transformer def}, in particular Remark \ref{rmk:assumptions for our model}.

A technical note is that we use $\infty$ in our constructions throughout this paper. In fact, we take $\infty$ to be the max possible value (with precision $O(\log(n + r(n)))$. More specifically, we can take the precision to be $C \cdot \log(n + r(n))$ for suitably large $C$, so that the max possible value is \textit{much} bigger than all other values needed to encode matrix and vector values.


\section{A Hierarchy Theorem for \CoT} \label{section:main results}

In this section, we state and prove the main results. Proofs in this section require a few primitives for manipulating transformers, which are provided in Appendix~\ref{sec:manipulate}. We start with the following lemma. Note that for padding to work (in general), $r(n) \geq n$ is a technical requirement. Padding as a technique is not translatable to the sub-linear regime.

\begin{lemma} \label{CoTAmplification}
    Let $\epsilon > 0$. Let $r(n) \geq n$ such that $r(n)$ is time-constructible, nondecreasing, unbounded, and $\log(r(n+1)) = (1+o(1)) \log(r(n))$. If $\CoT(r(n)) = \CoT(r(n)^{1 + \epsilon})$, then $\CoT(r(n)^{1 + \epsilon}) = \CoT(r(n)^{1 + 2 \epsilon})$. 
\end{lemma}

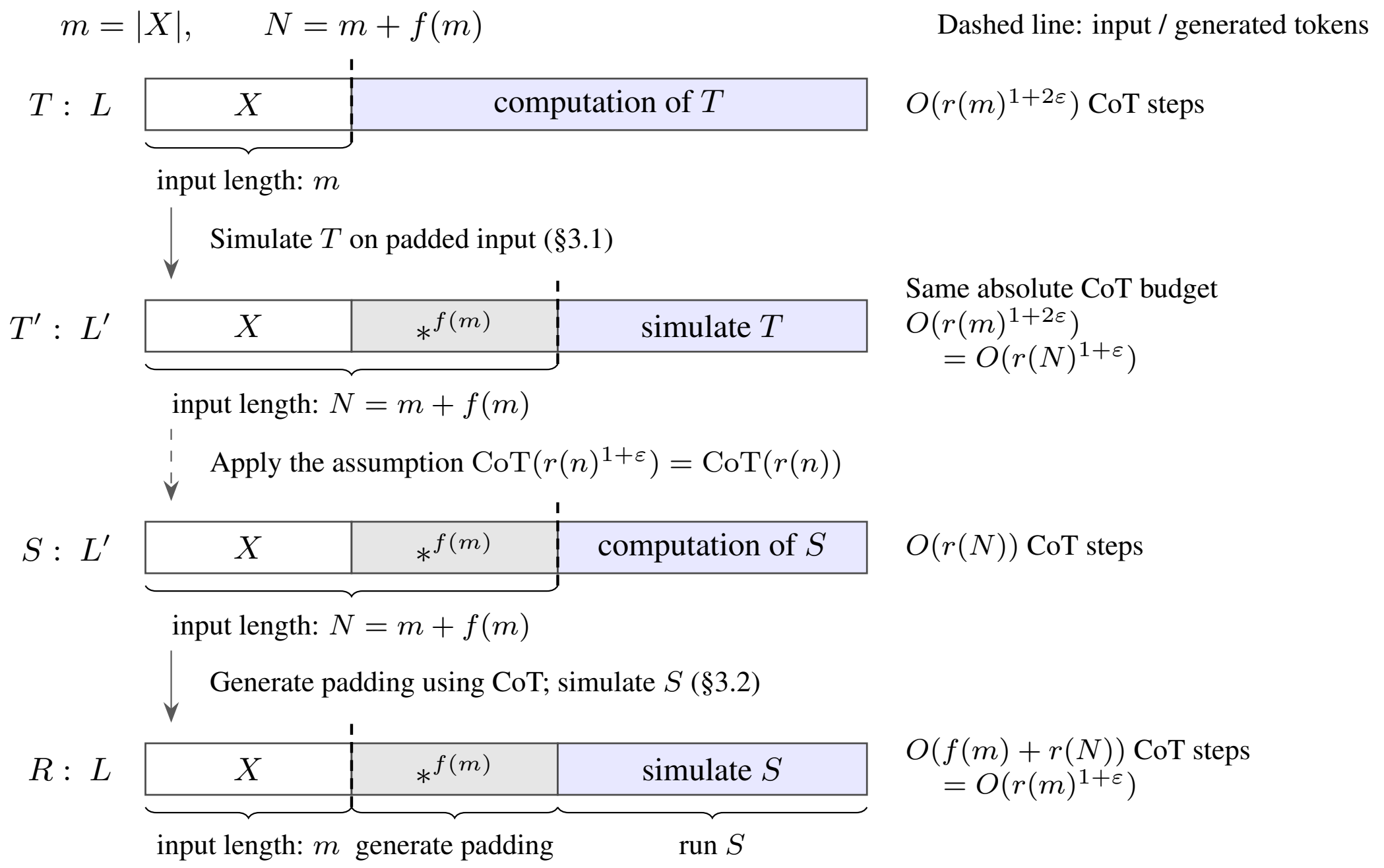
\begin{figure}[tbp]
  \centering
  \resizebox{\linewidth}{!}{
\begin{tikzpicture}[x=1cm,y=1cm,font=\small,
  segment/.style={draw=black!70,line width=.5pt},
  boundary/.style={densely dashed,thick},
  transition/.style={-{Stealth[length=2mm]},black!65},
  brace/.style={decorate,decoration={brace,mirror,amplitude=3pt}},
  note/.style={font=\footnotesize,align=left}]
\node[anchor=west] at (-.95,1) {$m=|X|,\qquad N=m+f(m)$};
\node[anchor=east,font=\footnotesize] at (12,1) {Dashed line: input / generated tokens};
\foreach \y/\lab in {0/{T:\ L},-2.15/{T':\ L'},-4.3/{S:\ L'},-6.45/{R:\ L}} {
  \node[anchor=east] at (-.2,\y+.25) {$\lab$};
  \draw[segment,fill=white] (0,\y) rectangle (2,\y+.5);
  \node at (1,\y+.25) {$X$};
}
\draw[segment,fill=blue!9] (2,0) rectangle (7,.5);
\node at (4.5,.25) {computation of $T$};
\draw[boundary] (2,-.12)--(2,.72);
\draw[brace] (0,-.12)--(2,-.12) node[midway,below=4pt,font=\footnotesize] {input length: $m$};
\node[note,anchor=west] at (7.25,.25) {$O(r(m)^{1+2\varepsilon})$ CoT steps};
\foreach \y in {-2.15,-4.3,-6.45} {
  \draw[segment,fill=black!10] (2,\y) rectangle (4,\y+.5);
  \node at (3,\y+.25) {$*^{f(m)}$};
  \draw[segment,fill=blue!9] (4,\y) rectangle (7,\y+.5);
}
\node at (5.5,-1.9) {simulate $T$};
\node at (5.5,-4.05) {computation of $S$};
\node at (5.5,-6.2) {simulate $S$};
\foreach \y in {-2.15,-4.3} {
  \draw[boundary] (4,\y-.12)--(4,\y+.72);
  \draw[brace] (0,\y-.12)--(4,\y-.12) node[midway,below=4pt,font=\footnotesize] {input length: $N=m+f(m)$};
}
\node[note,anchor=west] at (7.25,-1.9) {Same absolute CoT budget\\$O(r(m)^{1+2\varepsilon})$\\$\quad{}=O(r(N)^{1+\varepsilon})$};
\node[note,anchor=west] at (7.25,-4.05) {$O(r(N))$ CoT steps};
\draw[boundary] (2,-6.57)--(2,-5.73);
\draw[brace] (0,-6.57)--(2,-6.57) node[midway,below=4pt,font=\footnotesize] {input length: $m$};
\draw[brace] (2,-6.57)--(4,-6.57) node[midway,below=4pt,font=\footnotesize] {generate padding};
\draw[brace] (4,-6.57)--(7,-6.57) node[midway,below=4pt,font=\footnotesize] {run $S$};
\node[note,anchor=west] at (7.25,-6.2) {$O(f(m)+r(N))$ CoT steps\\$\quad{}=O(r(m)^{1+\varepsilon})$};
\draw[transition] (.25,-.75)--(.25,-1.45);
\node[note,anchor=west] at (.5,-1.08) {Simulate $T$ on padded input (\S3.1)};
\draw[transition,dashed] (.25,-2.9)--(.25,-3.6);
\node[note,anchor=west] at (.5,-3.23) {Apply the assumption $\mathrm{CoT}(r(n)^{1+\varepsilon})=\mathrm{CoT}(r(n))$};
\draw[transition] (.25,-5.05)--(.25,-5.75);
\node[note,anchor=west] at (.5,-5.38) {Generate padding using CoT; simulate $S$ (\S3.2)};
\end{tikzpicture}}
  \caption{Padding argument for Lemma \ref{CoTAmplification}.
  Write $m=|X|$ and $N=m+f(m)$, with $f$ chosen as in the proof.
  The transformer $T'$ simulates $T$ on a padded input with
  essentially the same absolute CoT budget.
  The assumed class equality yields a transformer $S$ with
  budget $O(r(N))$. Finally, $R$ generates the padding using
  CoT and simulates $S$.
  Dashed vertical lines separate supplied input from generated
  tokens; gray blocks denote padding. Bar lengths are schematic.}
  \label{fig:padding-outline}
\end{figure}

\begin{proof}[Proof of Lemma \ref{CoTAmplification}] First, we give an outline of how padding works (see Figure \ref{fig:padding-outline} for an illustration). Let $L$ be a language in $\CoT(r(n)^{1 + 2 \epsilon})$. We will show that $L \in \CoT(r(n)^{1 + \epsilon})$ under the assumption that $\CoT(r(n)) = \CoT(r(n)^{1 + \epsilon})$. Define a ``padded'' language $L' = \{X (*)^{f(|X|)} : X \in L\}$ (for an appropriately chosen function $f$), where $X (*)^{f(|X|)}$ means a string $X$ padded with $f(|X|)$ many $*$'s. Note that $L$ and $L'$ are essentially the same except the input is padded with $*$'s. Thus, using the transformer $T$ for $L$ as a black box, we design a transformer $T'$ for $L'$ using the same absolute number of \CoT\ steps, i.e. the number of \CoT\ steps of $T$ on input $X$ is essentially the same as the number of \CoT\ steps of $T'$ on input $X*^{f(|X|)}$.  However, as a function of input length, $T'$ will recognize $L'$ in $r(n)^{1 + \epsilon}$ \CoT\ steps. Now, by the assumption that  $\CoT(r(n)) = \CoT(r(n)^{1 + \epsilon})$, we obtain that $L'$ is indeed in $\CoT(r(n))$ via a transformer $S$. Since $L$ and $L'$ are essentially the same languages, we can design a new transformer for $L$ by padding an input $X$ appropriately and simulate $S$ on the padded input. This ensures that the new transformer for $L$ runs in $r(n)^{1+\epsilon}$ $\CoT$ steps. 



While this is the high-level approach, there are several technical challenges to adapt this to the transformer setting. First, to simulate $T$, one technical challenge is that $T$ expects an unpadded input over $\Sigma$. If we just run $T$ as it is on $X *^{f(|X|)}$, then since attention of CoT steps are over $*$ tokens as well, this would change the output for $T$ in ways that we can't control. Another challenge is to copy the output of a \textit{position} $i$ to a position $j > i$ far apart (where position $i$ means the $i^{\text{th}}$ token in the input string) -- so that desired calculations can be carried out. 
To simulate $S$, a technical challenge is when we receive $X$ as input, we need to include layers and heads that pads * tokens using CoT steps. The difficulty is to make sure that these layers/heads take effect only in certain steps.   

We now give details of the construction. 
\textit{In this proof, we will focus on a polynomial $r(n)$}. We will explain how this can be extended to other classes of functions after the proof. 

Assume $\CoT(r(n)) = \CoT(r(n)^{1 + \epsilon})$. 
Let  $L \in \CoT({r(n)}^{1 + 2\epsilon})$, and let $T$ be a transformer that recognizes $L$ in $O(r(n)^{1 + 2\epsilon})$ CoT steps. Let $\Sigma$ be the alphabet for $L$.

Define the padded language  
$L' = \{X (*)^{\floor{|X|^c}} : X \in L, * \not\in \Sigma\}$, where $c = \frac{1 + 2\epsilon}{1 + \epsilon}$ and $*$ is a new character (just used for padding). So, the alphabet for $L'$ is $\Sigma' = \Sigma \cup \{*\}$. 

As mentioned above, the proof consists of the constructions of the following two transformers. 

\noindent\textbf{\underline{Transformer $T'$}}: Using transformer $T$, we can construct transformer $T'$ that recognizes $L'$ in $O(r(n)^{1 + \epsilon})$ $\CoT$ steps. Then, by our assumption, $L' \in \CoT(r(n))$. Let $S$ be the transformer that recognizes $L'$ in $O(r(n))$ CoT steps. 

\noindent\textbf{\underline{Transformer $R$}}: Using transformer $S$, we construct a transformer $R$ that recognizes $L$ in $O(r(n)^{1 + \epsilon})$ CoT steps.

\subsection{Construction of \texorpdfstring{$T'$}{T'}}
    
\textit{Note that for any $i \in \N$, we say $x_i$, $\x_i$, or $i$ to refer to position $i$, i.e. step number $i$, that a transformer takes.}
    
We first give a high-level description of our construction: We take as input strings that look like $X = x_1 \dots x_m x_{m+1} \dots x_{m+\floor{m^c}}$, where $m \in \N$. Let $X_1 := x_1 \dots x_m$ and $X_2 := x_{m+1} \dots x_{m+\floor{m^c}}$. Let $\x_i$ denote the embedding of token $x_i$, $\x_i^{(l)}$ denote the intermediate value after layer $l$, and $\y_i$ denote the output at token $x_i$. Our goal is to design a transformer to check if $X \in L'$, i.e. check if $X_1 \in L$ and $X_2 = *^{\floor{m^c}}$. To check if $X_1 \in L$, we use a mask function to ``ignore" tokens in $X_2$, and run CoT steps as in $T$ by attending over tokens in $X_1$ and CoT-generated tokens $\x_{m + \floor{m^c} + 1}, \x_{m + \floor{m^c} + 2}, \dots$ \textit{Note that the mask function is not necessary in our construction, it just helps with making our proof concise -- see Remark \ref{rmk:simulate attn mask}.} The challenge is to make $\x_{m + \floor{m^c} + 1}$ equal to $\y_{m}$ so that the first CoT step of $T'$ is exactly the same as the first CoT step of $T$. To check if $X_2 = *^{\floor{m^c}}$, we just need a single decoding step.  

Let the parameters of $T$ be: alphabet: $\Sigma$; embedding dimension: $d$; number of layers: $L$; number of heads per layer: $H$; output function: affine linear transformation $\gamma_T$. 

We construct $T'$ with parameters: alphabet: $\Sigma' = \Sigma \cup \{*\}$; input string: $x_1 x_2 \dots x_{n}$, where $n = m + \floor{m^c}$, $m \in \N$; embedding dimension: $D = 2d + 22$. The embedding dimensions are (note that more details will be provided later in the proof): 
\begin{itemize}[noitemsep,leftmargin=10pt]
    \item All WE and PE used for $L$ remain the same. Say they occupy dimension numbers $1$ to $d$ of our vectors. When simulating $T$, $T'$ acts on these dimensions only. 
    \item 1 dimension for saving is $x_i = *$, WLOG, say dimension number $d+1$. 
    \item 5 dimensions for additional PE: $i, m, m+\floor{m^c}, \frac{1}{\floor{m^c}}, \frac{1}{i}$. 
    \item Dimensions that serve as placeholders. They are: 
    \begin{itemize}[noitemsep,leftmargin=10pt]
        \item 1 dimension for saving if $x_{m+1} = \dots = x_{m + \lfloor m^c\rfloor} = *$, say dimension $d+2$.
        \item 1 dimension for saving $\1(\Att = 1/\floor{m^c})$, say dimension $d+3$.
        \item $d$ dimensions for saving the copied $\x_\eta^{(L+1)}$.
        \item Other placeholder dimensions.
    \end{itemize}
\end{itemize}

{\em Word embedding (WE):} We define WE for $T'$ by using the exact same WE as for $T$, except for adding a new dimension (dimension number $d+1$) which is $1$ iff character is $*$ (and $0$ otherwise). That is, $\WE(*)$ is $1$ at dimension $d+1$, and $0$ elsewhere.

{\em Positional encoding (PE):} In addition to PE of $T$, we also use a supplementary PE (so that in case $T$ changes the values of its original PE via calculations, we still have additional PE values to use). For input length $n = m + \floor{m^c}$, we define the supplement PE to be $\PE_n(i) = \begin{bmatrix}
            i &m & m+\floor{m^c} &\frac{1}{\floor{m^c}} &\frac{1}{i}
        \end{bmatrix}$. Note that $\PE_n : \N \to \R^D$. We just write the 5 dimensions that matter (the rest are understood to be all $0$'s). We save supplement PE in 5 unique dimensions in embedding vectors (i.e. $\x_i$'s).

        {\em Heads and layers:} $T'$ has $L' := L+5$ layers, each layer has $H$ heads.  
        \begin{itemize}[noitemsep,leftmargin=10pt,topsep=2pt]
            \item Layers $1$ to $L$: exactly the same as the layers of $T$. Note that each head in these layers acts on dimensions $1$ to $d$ (i.e. the dimensions that $T'$ shares with $T$) only.
            \item Layers $L+1$ to $L+5$: additional layers, to be described.
        \end{itemize}

        {\em Multi-layernorms:} put multi-layernorms only where $T$ has.

        {\em Attention mask function:} we use causal mask; the only difference is that 
        \begin{equation*}
            \mask(i, j) = - \infty \qquad\text{if $i > m + \floor{m^c}$ and $j \in \{m+1, m+2, \dots, m+\floor{m^c}\}$.}
        \end{equation*}
        Note again that this tailored attention mask is avoidable (see Remark \ref{rmk:simulate attn mask}). That is, causal mask suffices; the use of the tailored mask is just for conciseness.

        {\em Output function:} $\gamma_{T'} := \delta_2 \circ \gamma_T \circ \delta_1$, where $\delta_1, \delta_2$ are affine linear transformations $\R^d \to \R^d$. Since composition of affine linear transformations is an affine linear transformation, we have that $\gamma_{T'}$ is an affine linear transformation.    

{\bf \em How $T'$ runs: } 
    \begin{itemize}[noitemsep, leftmargin=10pt, topsep=0pt]
        \item For $x_1$ to $x_m$: Run exactly as $T$ would run.

        Now, we denote the vector at position $m$ after layer $L$ (i.e. the last layer) of $T$ be $\v_T$. Note that we have $\x_m^{(L)} = \v_T$ (since $T'$ runs exactly as $T$ for its first $L$ layers). 

        \textit{Note that $\x_m^{(L)}$ has dimension $D$, while $\v_T$ has dimension $d$. When we write $\x_m^{(L)} = \v_T$, it actually means that ``at the dimensions that $T'$ shares with $T$, we have $\x_m^{(L)} = \v_T$."}

        \item  For $x_{m+1}$ to $x_{m + \floor{m^c} -1}$: It does not matter what $T'$ does here.  
        
        \item For $x_{m + \floor{m^c}}$: Copy the first $d$ dimensions of $\x_m^{(L)} (=\v_T)$ to this position. Also check if $x_{m+1} = x_{m+2} = \dots = x_{m+\floor{m^c}} = *$. 

        First, we add a layer (layer L+1) \textit{with no head}, so as to use its feedforward network to calculate $\1(j = m)$ and $\1(j = m + \floor{m^c})$ (as in Lemma \ref{lem:tran checks if c_i = t}). Note that layer needs to share the number of heads with other layers, but we can just make all heads ``silent."
        
        \textit{Note that for each $\1$ function, we need $3$ placeholder dimensions and $1$ dimension. In total, we need $8$ dimensions.} 

        To copy $\x_m^{(L)}$ to position $m + \floor{m^c}$, we do the following.
        \begin{itemize}[noitemsep, leftmargin=10pt, topsep=0pt]
            \item We add a layer (layer $L+2$). This layer has a head (head $1$). For query $i$ and key $j$, choose $\W_q$ and $\W_k$ such that 
            \begin{equation*}
                \q_i^T \k_j = 1 \cdot \big(-\frac{1}{j}\big) + \1(i = m + \floor{m^c}) \cdot \1(j = m), 
            \end{equation*}
            where $\1(i = m + \floor{m^c})$ and $\1(j = m)$ are calculated from layer $L+1$; $1$ can be easily obtained by adding a $1$ in a designated dimension in any layer before $L+2$. 
            
            We just use \textit{2 additional dimensions}, one to line up $1$ for $\q_i$ and $\big(-\frac{1}{j}\big)$ for $\k_j$, and the other to line up  $\1(i = m + \floor{m^c})$ for $\q_i$ and $\1(j=m)$ for $\k_j$.

            Note that when average hardmax is applied: 
            \begin{equation*}
                \eta = \eta(i) :=\argmax_{j\,:\,j \leq i} (\q_i^T \k_j) = \begin{cases}
                    i &\text{if $i \neq m + \floor{m^c}$} \\
                    m &\text{if $i = m + \floor{m^c}$}.
                \end{cases}
            \end{equation*}
            We design $\W_v$ such that $\W_v$ copies dimensions $1$ to $d$ of $\x_\eta^{(L+1)}$ to some $d$ placeholder dimensions of $\x_i^{(L+2)}$.

            \textit{We need $d$ placeholder dimensions here.}

            \item Then, at output function $\gamma_{T'}$, before applying $\gamma_T$, we perform a linear transformation to switch the $d$ placeholder dimensions above to dimension $1$ to $d$ using the linear transformation $\W$. Note that only at $i = m + \floor{m^c}$ do changes appear: $\v_T$ is now ``copied" to position  $m+\floor{m^c}$ of $T'$, and we obtain $\y_{m+ \floor{m^c}} = \gamma_T(\v_T)$. This makes $\x_{m+\floor{m^c} + 1}$ in $T'$ equal to $\x_{m+1}$ in $T$, as desired.
        \end{itemize}

        To check if $x_{m+1} = \dots = x_{m+\floor{m^c}} = *$, we need 3 additional layers (layers $L+3, L+4, L+5$). Say we want to save this in dimension number $d+2$.
        \begin{itemize}
            \item Design the following head (head 1 of layer $L+3$). Attend to all positions where character is *. Make query to be just $m$ (or any positive constant available), and make keys to be $\1(x_j = *)$ (by simply taking the value at the dimension $d+1$). Make values to be all $\1(j = m+\floor{m^c})$. We have the following cases: 
            \begin{itemize}
                \item If $x_{m+1} = \dots = x_{m + \floor{m^c}} = *$, then attention weights are exactly $1/\floor{m^c}$ at positions $m+1, \dots, m+\floor{m^c}$ and $0$ elsewhere. Therefore, attention (Att) is $1/\floor{m^c}$ (as values are $0$ except for being $1$ at position $m + \lfloor m^c \rfloor$). 
                \item If among $x_{m+1}, \dots, x_{m+\floor{m^c}}$, some are $*$ and the others are not, then Att is either $0$ or $1/t$ (where $t < \floor{m^c}$). 
                \item  If among $x_{m+1}, \dots, x_{m+\floor{m^c}}$, none are $*$, then we have uniform attention over all positions $1, \dots, m + \floor{m^c}$. So, Att is $\frac{1}{m + \floor{m^c}} \cdot 1 = \frac{1}{m + \floor{m^c}}$. 
            \end{itemize}

            \item Now, we just need to check if $\Att = \frac{1}{\floor{m^c}}$, where $\frac{1}{\floor{m^c}}$ is given from PE. By Lemma \ref{lem: check if 2 reals equal}, we use the feedforward network of layer $L+3$ and layer $L+4$ to get $\1(\Att = \frac{1}{\floor{m^c}})$, as desired. Say the result is saved in dimension $d+3$.

            \textit{Note that we need 3 additional dimensions to use the construction in Lemma \ref{lem: check if 2 reals equal}.} 

            \item Last, we use a layer (layer $L+5$) to implement the feedforward network as in Lemma $\ref{lem:if clause for non-neg entries}$, whose effect is to copy the results in dimension $d+3$ to dimension $d+2$ iff $\1(i = m + \floor{m^c})$. 

            \textit{Note that we need $3$ additional dimensions for this.}
        \end{itemize}
        
        \item CoT steps: Run exactly as if $T$ would run (on the dimensions that $T$ acts on, ignoring other dimensions). Attention mask makes each CoT step attend to only $x_1$ to $x_m$ and other CoT steps (i.e. $\x_{m+\floor{m^c}+1}, \x_{m+\floor{m^c}+2}, \dots$, which are equivalent to $\x_{m+1}, \x_{m+2}, \dots$ of $T$, respectively). 
        
        Let the transformer run $O(r(m)^{1 + 2 \epsilon})$ CoT steps (as $T$ does). 

        \item Last, we output True (i.e. $x_1 \dots x_{m + \floor{m^c}}$ is indeed in $L'$) if the following are correct: 
        \begin{itemize}
            \item The result of $T'$ (equivalently $T$) on $x_1 x_2 \dots x_m$ says that $X_1 = x_1 x_2 \dots x_m$ is in $L$. 
            \item $\1(x_{m+1} = \dots = x_{m+ \lfloor m^c \rfloor} = * ) = 1$. 
        \end{itemize}
    \end{itemize}
    This concludes our construction for $T'$. 

    We obtain that $L'$ can be recognized in $O(r(m)^{1+2\epsilon})$ CoT steps by the transformer $T'$. As $r$ is of polynomial order and $c = \frac{1 + 2\epsilon}{1 + \epsilon}$, we have  
    \begin{equation*}
        O(r(m)^{1+ 2\epsilon}) = O(r(m + \floor{m^{\frac{1+2\epsilon}{1 + \epsilon}}})^{1+\epsilon}). 
    \end{equation*}
    As we consider the input of $L'$ is of length $m + \floor{m^c}$, we obtain that $L' \in \CoT(r(n)^{1 + \epsilon})$, as desired. 

    

\begin{remark} \label{rmk:simulate attn mask}
    We can simulate the effects of the attention mask for $T'$, as detailed below. 
 	\begin{itemize}
 		\item Prepend five first layers of the transformer to do the following: 
 		\begin{itemize}
 			\item Layers 1, 2: Calculate $\1(i \geq m + \lfloor m^c\rfloor + 1)$ by Corollary \ref{cor:check t_1 <= x <= t-2}. 
 			\item Layers 3, 4: Calculate $\1(m+1 \leq i \leq m + \lfloor m^c\rfloor)$ by Corollary \ref{cor:check t_1 <= x <= t-2}. 
 			\item Layer 5: Use feedforward network to multiply $- \infty$ to $\1(m+1 \leq i \leq m + \lfloor m^c\rfloor)$ to get $(- \infty) \1(m+1 \leq i \leq m + \lfloor m^c\rfloor)$.  
 		\end{itemize}
 		\item Next, for every layer of $T'$ \textit{that does not involve multi-layernorm}, use the exact same heads as for the original $T'$ with the only difference that for every head, for any query $i$ and key $j$, embed the query $\q_i$ and key $\k_j$ such that: 
 		$$\q_i^T \k_j = \1(i \geq m + \lfloor m^c\rfloor + 1) \big(\1(m+1 \leq j \leq m + \lfloor m^c\rfloor) \cdot (-\infty)\big) + S,$$
        where $S$ is the value $\q_i^T \k_j$ in the original $T'$.
        
 		So, when $i \geq m + \lfloor m^c\rfloor + 1$ and $m+1 \leq j \leq m + \lfloor m^c\rfloor$, we have $\q_i^T \k_j = -\infty$, 
        meaning query $i$ will not attend to key $j$. Otherwise, $\q_i^T \k_j = 0 + S = S$, so attention is calculated as normal. 
 		
 		\item On the other hand, for every layer of $T'$ \textit{that involves multi-layernorm}, multi-layernorm will alter the values $v_{i,1} := \1(i \geq m + \lfloor m^c\rfloor + 1)$ and $v_{i,2} := \1(m+1 \leq i \leq m + \lfloor m^c\rfloor)$, which is not what we want.  
 		
 		To solve this, say $\begin{bmatrix}
 			\x_i &v_{i,1} &v_{i,2}
 		\end{bmatrix}$ is the current vector, and $T'$ wants to perform multi-layernorm on the vector $\x_i$ only. Denote $\bar{\x_i} = \multilayernorm(\x_i)$. We can add a layer just to perform multi-layernorm on $\x_i$ (by making a head such that value at $j$ is $\bar{\x_j}$, key at $j$ is $\frac{j - \text{ mean}}{\text{norm}}$, and query is $1$ -- this head gives $\bar{\x_i}$ as attention). We save $\bar{\x_i}$ into placeholder dimensions. So, after this layer, we yield $\begin{bmatrix}
	 		\x_i &v_{i,1} &v_{i,2} &\bar{\x_i}
	 	\end{bmatrix}$. 
 	
 		Next we add a layer on what $T'$ would have performed on $\bar{\x_i}$. The difference is that now we already have $\bar{\x_i}$, so no need to do multi-layernorm anymore (therefore, the numbers $v_{i,1}$ and $v_{i,2}$ are intact). We just need to use the numbers $v_{i,1}$ and $v_{i,2}$ to reproduce the mask and perform calculations just on $\bar{\x_i}$ (like in the bullet point above).

 		Thus, we have obtained the desired mask effect. 
 	\end{itemize} 
\end{remark}

\subsection{\texorpdfstring{Construction of $R$}{Construction of R}} 
    We have shown that $L' \in \CoT(r(n)^{1+\varepsilon})$. By the assumption that $\CoT(r(n)) = \CoT(r(n)^{1+\varepsilon})$, we obtain  $L' \in \CoT(r(n))$. Let $S$ be the transformer that recognizes $L'$ in $O(r(n))$ CoT steps.  We will use $S$ to construct a transformer $R$ that recognizes $L$ in $O(r(n)^{1+\epsilon})$ \CoT\ steps. 

    On a high level, we design $R$ as follows. $R$ gets $X = x_1 \dots x_m$ as input. On $X$, we want to simulate $S$. To do that, for CoT steps $\leq \lfloor m^c \rfloor$ (i.e. at positions $m+1, \dots, m + \lfloor m^c \rfloor$), we generate letters $*$ (i.e. we are trying to reconstruct $X (*)^{\lfloor m^c\rfloor}$). Then, on CoT steps $> \lfloor m^c\rfloor$, we simulate $S$ (as if we are in the CoT steps of $S$ on $L'$).  

    Let the parameters of $S$ be: embedding dimension: $d$; number of layers: $L$; number of heads per layer: $H$; output function: $\gamma_S$. 

   We now describe the parameters of $R$: 
        
        {\em Input} (after embedding as for $S$): $\x_1, \x_2, \dots, \x_m$, where $\x_i \in \R^d$.  


        {\em Positional encoding:} as noted in Remark \ref{rmk:assumptions for our model}: 
        \begin{itemize}[noitemsep,leftmargin=10pt,topsep=0pt]
            \item For $\x_i$, save $i$ to encode position. Also, use fixed numbers $m, \floor{m^c}$ in PE. So, $\PE_m(i) = \begin{bmatrix}
                i &m &\floor{m^c}
            \end{bmatrix}$
            \item For CoT steps: at $\x_i$ where $i > m$, PE is updated to $i$. 
        \end{itemize}

        {\em Embedding dimension:} $D = 3d+4$, namely 
        \begin{itemize} [noitemsep,leftmargin=10pt,topsep=0pt]
            \item $d$ for $\x_i$ (matching with $S$). WLOG, suppose: 
            \begin{itemize}
                \item S acts on dimension $1$ to $d$. 
                \item Dimension $d$ is to save $\1(x_i = *)$. 
            \end{itemize}
            \item Some 3 dimensions for positional encoding.
            \item $2d+1$ additional dimensions (WLOG, dimensions $d+1$ to $3d+1$). In particular: 
            \begin{itemize}[noitemsep,leftmargin=10pt]
                \item Dimensions $d+1$ to $3d+1$ is for saving $\bar{\v}$ (this will be explained below).
            \end{itemize}
        \end{itemize}
        
        {\em Number of layers and heads:} $L+3$ layers, $H$ heads in each layer. 
        \begin{itemize}[noitemsep, leftmargin=10pt,topsep=0pt]
            \item The first $L$ layers of $R$ match with $S$ (these layers act only on dimensions that $S$ has effects on (i.e. dimensions $1$ to $d$)). 
        \end{itemize}

        {\em Attention masks:} standard causal mask.

        {\em Output function:} $\gamma_R = \delta \circ \gamma_S$, where $\delta$ is an affine linear transformation. 

    {\em Descriptions and justifications of $R$: }
    \begin{itemize}[noitemsep,leftmargin=10pt,topsep=0pt]
        \item For $\x_1, \dots, \x_{m-1}$: Run exactly as $S$. 

        \item For $\x_m, \dots, \x_{m + \floor{m^c} -1}$ (i.e CoT steps $1$ to $\floor{m^c}$): 
        
        \begin{itemize}
            \item Run calculations exactly as for $S$. 
            \item Also, we need to simulate paddings here: for $\x_i$ ($m \leq i \leq m + \floor{m^c} - 1$), we want the output $\y_i$ of position $i$ to be $*$ (i.e. $\y_i$ has WE for $*$: $0$ from dimensions $1$ to $d-1$, and $1$ at dimension $d$). 
            \begin{itemize}[noitemsep,leftmargin=10pt]
                \item First, we use 2 layers (layers $L+1$ and $L+2$) to compute $f(i) := \1(m \leq i \leq m + \floor{m^c} - 1)$ as in Corollary \ref{cor:check t_1 <= x <= t-2}. We save $f(i)$ in a designated dimension. Also, we save $g(i)$ at the dimension to indicate $*$ (i.e. dimension $d$).    

                \item Let $\v = \v(i) :=\x_i^{(L+2)}$ represent the vector that we obtain for query $i$ after layer $L+2$, restricted on dimensions $1$ through $d-1$. The idea is that we would want to make $\v = \0_{d-1}$ if $f(i) = 1$, and $\v$ intact otherwise. We proceed as follows. 

                Use a previous feedforward network (say, of layer $L+2$) to save $\v$ in some separate $d-1$ placeholder dimensions, say, WLOG, dimensions $d+1$ to $2d-1$. 

                Next, use a layer (layer $L+3$) to calculate $\bar{\v} := f(i) \v$ (on dimensions $d+1$ to $2d-1$) - this can be done by Lemma \ref{lem:mult vect by 0 or 1}.  

                Last, in the output function, calculate $\v - \bar{\v}$ (on dimensions $1$ to $d-1$). Note that this returns $\v$ if $f(i) = 0$, and $\0_{d-1}$ if $f(i) = 1$, as desired.

                \item Thus, the effect is that $x_{i} = *$, for all $m+1 \leq i \leq m + \floor{m^c}$. 
            \end{itemize}

            \textit{Note that in total, we need $3 + 2(d-1) = 2d+1$ additional dimensions.}
        \end{itemize}

        \item For later steps: for $\x_{m + \floor{m^c}}$ (and after): just run $S$. For instance, at $\x_{m + \floor{m^c}}$, it is exactly as if $S$ runs on the input $X' = x_1 \dots x_{m + \floor{m^c}}$. Subsequently, step $m + \floor{m^c} + r$ of $R$ is exactly the same as CoT step number $r$ of $S$. 

        Therefore, $R$ will be able to accept or reject $X = x_1 \dots x_m$ correctly in 
        $$(\floor{m^c}) + O(r(m + \floor{m^c})) = O(r(m)^c)$$
        CoT steps. The construction of $R$ is completed.      
    \end{itemize}
     
    Now, with $R$, we obtain that $L \in \CoT(r(n)^c) = \CoT(r(n)^{\frac{1 + 2\epsilon}{1 + \epsilon}}) \subseteq \CoT(r(n)^{1 + \epsilon})$. In particular, we obtain $L \in \CoT(r(n)^{1 + \epsilon})$. 
    
    Thus, we can conclude that $\CoT(r(n)^{1 + \epsilon}) = \CoT(r(n)^{1 + 2\epsilon})$. 
\end{proof}
This completes the proof of Lemma~\ref{CoTAmplification}.

\begin{remark}[Extending to time-constructible functions] 

Fix \(\varepsilon>0\), and suppose that \(r:\mathbb N\to\mathbb N\) is nondecreasing and unbounded, with $r(n) \geq n$ and
\[
\log r(n+1)=(1+o(1))\log r(n).
\]
The proof above will go through if we pad the input $X$ to $X (*)^{f(|X|)}$ for some function $f$, such that $r^{\frac{1+2\varepsilon}{1+\varepsilon}}(n) \leq r(n+f(n)) \leq r^{1+\varepsilon}(n)$. 

Write
\[
\alpha=\frac{1+2\varepsilon}{1+\varepsilon},
\qquad
\beta=1+\varepsilon,
\]
and choose a rational constant \(\theta\) satisfying \(\alpha<\theta<\beta\). Such a choice exists because
\[
\beta-\alpha=\frac{\varepsilon^2}{1+\varepsilon}>0.
\]
Define
\[
m(n)=\min\{m\in\mathbb N:m\ge n,\ r(m)\ge r(n)^\theta\},
\qquad
f(n)=m(n)-n.
\]
Unboundedness of \(r\) ensures that \(m(n)\) exists, and \(f(n)\) is a nonnegative integer. For all sufficiently large \(n\), we have \(r(n)>1\), so \(m(n)>n\), and minimality gives
\[
r(m(n)-1)<r(n)^\theta\le r(m(n)).
\]
Furthermore, the assumed growth condition and \(\beta/\theta>1\) imply that
\[
r(k+1)\le r(k)^{\beta/\theta}
\]
for all sufficiently large \(k\). Applying this with \(k=m(n)-1\ge n\), we obtain
\[
r(n)^\alpha
\le r(n)^\theta
\le r(n+f(n))
\le r(m(n)-1)^{\beta/\theta}
<r(n)^\beta.
\]
Thus padding \(X\) to \(X(*)^{f(|X|)}\) satisfies the required growth inequalities for all sufficiently large input lengths.
\end{remark}

\subsection{Proof of Main Theorem}

Equipped with Lemma~\ref{CoTAmplification}, we are ready to prove our main theorem. 

\begin{theorem} \label{thm:main theorem}
For $\epsilon > 0$ and $r(n) \geq n$ such that $r(n)$ is time-constructible, nondecreasing, unbounded, and $\log(r(n+1)) = (1+o(1)) \log(r(n))$, we have $\CoT(r(n)) \subsetneq \CoT(r(n)^{1+\epsilon}).$ 
\end{theorem}

\begin{proof}
    Let $\epsilon > 0$ be arbitrary. Suppose, for the sake of contradiction, that $\CoT(r(n)) = \CoT(r(n)^{1+\epsilon})$.
    
    By Theorem \ref{thm:hierrachy TIME vs CoT in MS23}, we have
    \begin{align*}
        \CoT(r(n)) \subseteq \TIME(r(n)^3) \subseteq \CoT(r(n)^3) \subseteq \TIME(r(n)^9) \subseteq \CoT(r(n)^9),
    \end{align*}

By Time Hierarchy Theorem, $\TIME(r(n)^3)\subsetneq \TIME(r(n)^9)$. Thus, from the chain of containment,
$\CoT(r(n)) \subsetneq \CoT(r(n)^9)$. 

    Now, note that $\CoT(r(n)) = \CoT(r(n)^{1 + \epsilon})$ by assumption. Then, by Lemma \ref{CoTAmplification}, we obtain that $\CoT(r(n)) = \CoT(r(n)^{1 + 2\epsilon})$. Now, applying the lemma again, we get that $\CoT(r(n)) = \CoT(r(n)^{1 + 4\epsilon})$. After $i$ applications, we get $\CoT(r(n)) = \CoT(r(n)^{1 +2^i \epsilon})$. By choosing $i=\lceil \log_2 {\frac{8}{\epsilon}}\rceil$, we get that $\CoT(r(n)) = \CoT(r(n)^{9})$, contradiction. 

    Thus, we can conclude now that $\CoT(r(n)) \subsetneq \CoT(r(n)^{1 + \epsilon})$. 
\end{proof}

\section{Conclusions and Open Problems} 

We established a hierarchy theorem for chain-of-thought reasoning, identifying the number of decoding steps as a fundamental computational resource. Our results raise an important open question of obtaining a hierarchy theorem that is as tight as classical time and space hierarchies in the Turing machine model. In particular, can we show that
\[
\CoT(r(n)) \subsetneq \CoT(r(n)\,\mathrm{polylog}(r(n)))?
\]

Our proof relies on simulations of transformers with Turing machines combined with padding. 
The known simulations incur quadratic blow-up, and it is unclear whether near-linear time simulations of $\CoT(r(n))$ are possible. In particular, can we show
\[
\CoT(r(n)) \subseteq \mathrm{TIME}(r(n)^{1+o(1)})?
\]
Moreover, due to the use of the padding technique, we only get Theorem \ref{thm:main theorem} for $r(n) \geq n$. The case $r(n) < n$ is an open question for saturated-attention transformers. We expect a positive answer.

Finally, the full computational power of chain-of-thought reasoning is not yet well understood and is a promising research theme. 

\begin{ack}

N. V. Vinodchandran and Thanh Le were supported in part by NSF awards 2342244, 2413848, and a UNL Grand Challenges Grant.
A. Pavan was supported in part by NSF awards 2342245 and 2413849.
\end{ack}

\printbibliography

\appendix



\section{Transformer Definition} \label{sec:transformer def}

A transformer is a neural network and a function $\Sigma^* \to \Sigma$. In this paper, we work with \textit{decoder-only} transformer. We define the model and its components below. We then explain assumptions for our constructions. 

\begin{definition}
    A decoder-only transformer with $H$ heads, $L$ layers, model dimension $d$ consists of the following components:
    
    \begin{enumerate}
        \item \textbf{Embedding function:}
        \[
        e : \Sigma \times \mathbb{N} \to \R^d.
        \]
        where for $a \in \Sigma$, $t \in \N$, input length $n$, we define:  
        \begin{equation}
            e(a, t) = \WE(a) + \PE_n(t).
        \end{equation}
        $\WE, \PE_n$ are a word embedding function and a positional encoding function, respectively, where
        \begin{align*}
            &\text{WE} : \Sigma \to \R^d \\ 
            &\text{PE}_n : [n] \to \R^d \text{ for $n \in \N$}.
        \end{align*}
        In particular, positional encoding is dependent on the length of the input string (see \cite{Strobl_2024} for remarks on this). 
        
        \item \textbf{Attention heads:} 
        For each layer $\ell \in [L]$ and head $k \in [H]$:
        \begin{align*}
            s_k^\ell &: \R^d \times \R^d \to \R 
            \quad \text{(similarity / score function)}, \\
            v_k^\ell &: \R^d \to \R^d
            \quad \text{(value function)}.
        \end{align*}
    
        \item \textbf{Activation block (feedforward + skip-connection):}
        For each layer $\ell \in [L]$,
        \[
        f^\ell : (\R^d)^{H} \times \R^d \to \R^d \quad\text{(activation function)}.
        \]
    
        \item \textbf{Output function:}
        \[
        \gamma : \R^d \to \R^d.
        \]
    \end{enumerate}

\end{definition}
\begin{definition}  
    Given an input sequence $x = (x_1, \dots, x_n) \in \Sigma^n$. We define one decoding step $\Sigma^n \to \Sigma$ of a decoder-only transformer as follows. 
    
    \begin{enumerate}
        \item \textbf{Embeddings:} 
        \[
        \x_i^{(0)} = e(x_i, i) \quad \text{for } i \in [n].
        \]

        \item \textbf{Multihead Self-attention:} For each layer $\ell \in [L]$ and each position $i \in [n]$, we compute $H$ attention heads. For head $k \in [H]$, we compute:  
        \begin{equation*}
            \a_{i}^{\ell, k} = \sum_{j=1}^{c(i)} \frac{s_k^\ell(\x_i^{(\ell-1)}, \x_j^{(\ell - 1)})}{Z_i^{\ell, k}} \cdot v_k^\ell(\x_j^{(\ell - 1)}), \quad\text{where } Z_{i}^{\ell, k} = \sum_{j=1}^{c(i)} s_k^\ell(\x_i^{(\ell - 1)}, \x_{j}^{(\ell - 1)}),
        \end{equation*}
        where $c(i)$ is $i$ for causal attention. 

        We define $s = s_k^\ell$ and $v = v_k^\ell$ as follows (note that we drop $\ell, k$ for conciseness). $s$ has two affine linear transformations $\R^d \to \R^d$ parametrized by $(\W_q, \b_q)$ (for queries) and $(\W_k, \b_k)$ (for keys) as parameters. Likewise, $v$ has one affine linear transformation $(\W_v, \b_v)$ (for values) as parameter.  
    
        \begin{align}
            &s(\x_i, \x_j) = \exp\bigg(\frac{\q_i^T \mathbf{k}_j}{\sqrt{d}} + \mask(i,j)\bigg), \\
            &v(\x_j) = \W_v \bar{\x_j} + \b_v
        \end{align}
        where $\bar{\x}_i = \mathcal{N}_1(\x_i)$, where $\mathcal{N}_1 = \Ncal_1(\ell)$ is either a multi-layernorm (defined below) or an identity function,
        \begin{align}
            &\q_i = \W_q \bar{\x_i} + \b_q \\
            &\k_i = \W_k \bar{\x_i} + \b_k, 
        \end{align}
        and $\mask$ is the attention masking function (defined below).

        Attention mask function, denoted $\mask : \N \times \N \to \{-\infty, 0\}$, decides whether keys get attended to or not. $\mask$ takes a query and a key as inputs and returns $-\infty$ if the key is forbidden from the query. $\mask$ returns $0$ otherwise. For instance, attention mask function for (standard) causal masking is: 
        \begin{equation*}
            \mask(i,j) = \begin{cases}
                0 &\text{if $j \leq i$}, \\ 
                -\infty &\text{otherwise.}
            \end{cases}
        \end{equation*}
        Different mask functions are studied (for example, see \cite{Swietojanski23}).

        \item \textbf{Activation block:} 

        The activation function $f = f^\ell$ first aggregates attention heads' outputs, and then feeds the results through a feedforward network (with skip-connection).

        For each position $i$, $f$ takes as input attention heads' outputs $\a_{i,1}, \dots, \a_{i, H} \in \R^d$ and the previous layer value $\x_i = \x_i^{(\ell -1)}$.  

        First $f$ sums up outputs for heads, yielding \begin{equation}
            \a_i = \sum_{k = 1}^H \a_{i,k}.
        \end{equation}
        $\a_i$ is then passed through a linear transformation, and then a two layer feedforward (FFW) network (with a multi-layernorm (or identity), a non-linearity, and skip-connections in between) as follows 
        \begin{align}
            &\o_i =  \W_0 \a_i + \b_0 + \x_i \\ 
            &\bar{\o}_i = \mathcal{N}_2(\o_i) \\
            &\x'_i = \W_2 \,\ReLU(\W_1 \bar{\o}_i + \b_1) + \b_2 + \o_i.  
        \end{align} 
        Then, we get that $\x_i^{(\ell)} = \x'_i$.

        \item \textbf{Output:}
        The final output at position $i$ is
        \[
        \y_i = \gamma(\x_i^{(L)}), 
        \]
        where $\gamma$ is an affine linear transformation from $\R^d \to \R^d$, parametrized by $(\W, \b)$. 
    \end{enumerate}
\end{definition}

\begin{remark}
    With the definition above, parameters for a decoder-only transformer, denoted $T$, include:

    \begin{itemize}
        \item Embedding dimension ($d$), number of heads ($H$), number of layers ($L$). 
        \item Word embedding function $\WE$ and positional encoding functions $\PE_n$ (for $n \in \N$). 
        \item Attention mask function. Default to be causal mask.
        \item For each attention head: 3 affine linear transformations $(\W_q, \b_q), (\W_k, \b_k), (\W_v, \b_v)$, where $\W$ denotes the linear transformation, and $\b$ denotes the affine part. 
        \item For each layer: 2 multi-layernorms $\Ncal_1$ and $\Ncal_2$ (which can be identity); an affine linear transformation $(\W_0, \b_0)$; a 2-layer FFW network (parametrized by $(\W_1, \b_1)$ and $(\W_2, \b_2)$).
        \item For output function: an affine linear transformation $(\W, \b)$. 
    \end{itemize}
\end{remark}

\begin{remark}
    For \textit{saturated attention} (also known as \textit{ average-hard attention}), we modify multihead self-attention as follows 
    \begin{align}
        &\a_i^{\ell, k} = \sum_{j=1}^{c(i)} \alpha_{i,j}^{\ell, k} v_k^\ell(\x_j^{(\ell - 1)}),
    \end{align}
    where $(\alpha_{i,j})_{j=1}^{c(i)} = \hardmax(s(\x_i, \x_1), s(\x_i, \x_2), \dots, s(\x_i, \x_{c(i)}))$. We define $\hardmax$ to take as input a vector of real numbers, and return a probability distribution (of the same dimension) whose weights are equally distributed between the maximum positions, and $0$ elsewhere. 

    Based on current theoretical results, saturated attention appears to be computationally equivalent to softmax attention, in the sense that no separation between them is known (see Table 1 of \cite{Strobl_2024}). From a practical point of view, trained softmax systems usually exhibit attention concentrated almost equally on a few dominant positions. Thus, the work of \cite{merrill2022saturated} argues that average hardmax is a realistic approximation of how Transformers behave in practice. This motivates average hardmax as a reasonable and mathematically simple abstraction of softmax attention.
\end{remark}
    
\begin{definition}[Multi-layernorm, \cite{merrill2023expressive}]
    Layernorm is defined as 
    \begin{equation}
        \layernorm(\x) = \frac{\x'}{||\x'||},
    \end{equation}
    where $\x' = \x - \text{mean}(\x)$, where $\text{mean}(\x)$ computes the mean of vector $\x$. 

    Then, let $\M : \R^d \to \R^d$ be a parameter matrix that projects vectors from $\R^d$ to $\R^d$. Then projected pre-norm is defined as 
    \begin{equation}
        \text{proj\_layer\_norm}(\v, \M) = \layernorm(\M \v).
    \end{equation}
    If context is clear, we write $\text{proj\_layer\_norm}(\v)$.

    Next, let $d$ be divisible by $k \in \N$. For $1 \leq i \leq k$, let $\M_i : \R^d \to \R^{d/k}$ be a parameter projection matrix. let $\langle\cdot\rangle_{i=1}^k$ denote iterated concatenation. The multi-pre-norm of $\v \in \R^d$ is defined as 
    \begin{equation}
        \text{multi\_layer\_norm}(\v, \M_1, \dots \M_k) = \langle \text{proj\_pre\_norm}(v, M_i)\rangle_{i=1}^k.
    \end{equation}
    There is a result that says (compositions of) projected pre-norm can simulate multi-pre-norm. We use multi\_layer\_norm in our transformer model (we denote it by $\Ncal$).
\end{definition}

\begin{remark} \label{rmk:assumptions for our model}
    Our constructions of decoder-only transformers assume the following: 
    \begin{itemize}
        \item Positional encoding (PE):
        \begin{itemize}
            \item PE uses the number $i$ to encode position $i$. In addition, PE may contain other fixed numbers independent of $i$ (such as those dependent on the input length).  
    
            \item Without loss of generality (WLOG), we assume that our transformers have PE for generated tokens. That is, $\x_i$ (for any $i > n$ where $n$ is the input length) has $i$ as PE. 
    
            (This can be implemented easily by just using the output function $\gamma$ to add $1$ to PE $j$ (for $\x_j$). If $j \leq n-1$, then this does not change anything, as the output is not taken to be the next token. If $j \geq n$, then that makes the PE for $\x_{j+1}$ exactly $j+1$.)
    
            \item By Lemma \ref{lem:i gives 1/i easily}, WLOG, we can use $1/i$ as a positional encoding value for $\x_i$. 
        \end{itemize} 
    
        \item Attention: We use saturated attention.
        \item Precision: We assume precision as in \cite{merrill2023expressive}. That is, for input of length $n$ and $r(n)$ intermediate decoding steps, precision is $O(\log(n + r(n)))$.
        \item Attention mask: We allow slightly tailored attention mask. That is, we forbid some CoT steps from attending to a certain portion of past CoT steps. 
        
        Note that this is to help with writing concise proofs rather than a necessary condition.
        \item Attention head's dimension: We don't follow the exact definition of attentions in \cite{merrill2023expressive}, as they consider dimension $d/H$ for each head. But, their model and ours can simulate each other easily. 
        \item Layernorm: We allow the possibility of applying or not applying multi-layernorm. 
    \end{itemize}

    In particular, Theorem \ref{thm:hierrachy TIME vs CoT in MS23} still holds under our assumptions.
\end{remark}

\begin{remark}
    Attention mask is not required for our constructions. Yet, we use attention mask to simplify our construction. We explain how to ``simulate" attention mask in Remark \ref{rmk:simulate attn mask}.
\end{remark}

\section{Primitives for Manipulating Transformers}\label{sec:manipulate}

In this subsection we list and prove computations that we will perform with transformers in our main proofs. The following lemma is implicit in \cite{chiang2022overcoming}.  We write it in a form that we will use. 
It essentially says that transformers can check equality between integers and a given integer.

\begin{lemma} \label{lem:tran checks if c_i = t}
    Let current prefix be $\x_1 \x_2 \dots \x_c$, where each $\x_i \in \R^6$, $\x_i = \begin{bmatrix}c_i & t & 0 & 0 & 0 & 0\end{bmatrix}^T$, for $c_i, t \in \Z$. Then, there exists a decoder-only transformer with 1 layer (which has 1 head) that, for any $\x_i$, computes $\1(c_i = t)$ and saves into dimension $6$. That is, there exists a decoder-only transformer that returns $\begin{bmatrix}
        c_i &t &0 &0 &0 &\1(c_i = t)
    \end{bmatrix}^T$.    
\end{lemma}

\begin{proof}

    We design a transformer $T$ as follows: 
    \begin{itemize}
        \item Dimension: $6$; number of layers: $1$; number of heads per layer: $1$. 
        \item $\WE$, $\PE$: not applied, since inputs are already given to vector form. 
        \item For layer $1$: put $\Ncal_1 = \Ncal_2 = \identity$; put $\W_0^{1} = \I$ and $\b_0 = \0_{6 \times 1}$. We specify $(\W_1^1, \b_1^1)$ and $(\W_2^1, \b_2^1)$ below. 
        \item For the head in layer $1$: put $\W_v^{1,1} = \0_{6 \times 6}$ and $\b_v^{1,1} = \0_{6 \times 1}$. The effect of this is to make head output $\a_i^{1,1} = \0_{6 \times 1}$ for all $1 \leq i \leq c$. That means, we do not want this head to have any effects. 
        \item Output function: we put $\W = \I_{6 \times 6}$ and $\b = \0_{6 \times 1}$. 
    \end{itemize}
    
    Now, fix a position $i$, with input $\x_i^{(0)} = \begin{bmatrix}c_i & t & 0 & 0 & 0 & 0\end{bmatrix}^T$.  

    Consider layer $1$. The head produces $\a_i^{1,1} =  \0_{6 \times 1}$. Since there is only one head $\a_i^1 = \a_i^{1,1}$. Since we know we talk about layer $1$, we can drop the superscript $1$, and just write $\a_i = \0_{6 \times 1}$. Then, since $\W_0^1 = \I$ and $\b_0 = \0_{6 \times 1}$, we yield that $\o_i = \x_i^{(0)}$. Next, since $\Ncal_2 = \identity$, we have that $\bar{\o}_i = \x_i$. 

    Now, let 
    \begin{equation*}
        \W_1^1 = \begin{bmatrix}
            0 &0 &0 &0 &0 &0 \\
            0 &0 &0 &0 &0 &0 \\ 
            1 &-1 &0 &0 &0 &0 \\
            1 &-1 &0 &0 &0 &0 \\
            1 &-1 &0 &0 &0 &0 \\
            0 &0 &0 &0 &0 &0
        \end{bmatrix}, \qquad \b_1^1 = \begin{bmatrix}
            0 \\0 \\-1 \\0 \\1 \\-1
        \end{bmatrix}. 
    \end{equation*}
    We yield 
    \begin{equation*}
        \ReLU(\W_1^1 \x_i + \b_1) = \begin{bmatrix}
            0 \\0 \\\max\{0, c_i - t-1\} \\\max\{0, c_i - t\} \\\max\{0, c_i - t +1\} \\ 0 
        \end{bmatrix}.
    \end{equation*}

    Now, observe that since $c_i, t \in \Z$, we have 
    \begin{equation*}
        \1(c_i = t) = \max\{0, c_i - t - 1\} - 2\max\{0, c_i - t\} + \max\{0, c_i - t+1\}.
    \end{equation*}
    
    Therefore, we let 
    \begin{equation*}
        \W_2^1 = \begin{bmatrix}
            0 & 0 &0 &0 &0 &0 \\
            0 & 0 &0 &0 &0 &0 \\
            0 & 0 &0 &0 &0 &0  \\
            0 & 0 &0 &0 &0 &0 \\
            0 & 0 &0 &0 &0 &0 \\
            0 &0 &1 &-2 &1 &0
        \end{bmatrix}, \qquad \b_2^1 = \0_{6 \times 1}.
    \end{equation*}
    We obtain that $\x_i^{(1)} = \begin{bmatrix}
        c_i &t &0 &0 &0 &\1(c_i = t)
    \end{bmatrix}$, as desired. 
    
\end{proof}

\begin{remark}
    Some notes on how we use different components of transformers to build desired properties:
    \begin{itemize}
        \item Placeholder dimensions: we call dimensions $3$ to $5$ (informally) as placeholder dimensions. Their purpose is to save results of auxiliary calculations.  
        \item Dimensions to save results: dimension $6$ is a dimension to save results. 
        \item Effects of 2-layer FFW network: multiply by constants, add constants, perform ReLU, move numbers between dimensions.
        \item Effects of skip-connection: keep original values. That is, values $c_i$ and $t$ in dimensions $1$ and $2$ are kept intact in $\x_i^{(1)}$ thanks to skip connections. 
        \item For dimensions that we do not want to change, we can make our heads and linear transformations ignore them by ignoring respective rows/columns. Or, we can make values of the dimensions be $0$ during calculations, and then recover their original values using skip-connections.  
    \end{itemize}
\end{remark}


We obtain the following corollary for checking if an integer $x$ is between 2 integers $t_1$ and $t_2$. 

\begin{corollary} \label{cor:check t_1 <= x <= t-2}
    Say $\x = \begin{bmatrix}
        x &t_1 & t_2 
    \end{bmatrix}$ where $x \in \Z$ and $t_1, t_2 \in \Z$. There exists two nested 2-layer feedforward networks (with ReLU) that compute $\1(t_1 \leq x \leq t_2)$. 
\end{corollary}

\begin{proof}
    First, use an affine linear transformation and a ReLU (of the first feedforward network) to compute 
    $f(x) := \ReLU(t_1 -x) + \ReLU(x - t_2).$
    Note that $f(x) = 0$ iff $t_1 \leq x \leq t_2$. Next, use a second feedforward network to compute if $\1(f(x) = 0) \equiv \1(t_1 \leq x \leq t_2)$. 
\end{proof}

\begin{remark}
    We can implement the nested feedforward networks of the above corollary using a transformer with $2$ layers and $3$ additional dimensions ($2$ as placeholder dimensions, and $1$ as result-saving dimension).
\end{remark}

Next, we have the following lemma for checking if $x = t$, where $t \in \R$ is given, and $x \in \R$. Note that we state our result for feedforward networks, but similar to Lemma \ref{lem:tran checks if c_i = t}, we can transfer feedforward networks to transformers by making attention heads silent and using additional dimensions (to avoid effects of skip-connections).  

\begin{lemma} \label{lem: check if 2 reals equal}
    Consider $\x = \begin{bmatrix}
        x &t &0 &0 
    \end{bmatrix}$ where $x, t \in \R$. Then, there exists two nested 2-layer feedforward networks (with $\ReLU$) that return $\begin{bmatrix}
        \_ &\_ &\_ &\1(x = t)
    \end{bmatrix}$.
\end{lemma}
\begin{proof}
    First, use an affine linear transformation to get $x - t$ and $t -x $ (saved in dimensions $3$ and $4$). Then, use $\ReLU$ to get $\ReLU(x-t)$ and $\ReLU(t-x)$. Then use an affine linear transformation to get $\ReLU(x-t) + \ReLU(t-x) = |x-t|$ (saved in dimension $4$).  Next, the second 2-layer feedforward network, compute $\ReLU(1 - \infty |x-t|)$ (saved in dimension 5), where $\infty$ is a scalar. We get $1$ if $x = t$, and $0$ otherwise. 
\end{proof}

\begin{remark}
    We can implement the two nested feedforward networks as in Lemma \ref{lem: check if 2 reals equal} by using 2 layers of a transformer, with 2 placeholder dimensions and 1 result-saving dimension. 
\end{remark}

Next, we have the following lemma from \cite{perez2021attention}. We generalize so that it works for our case. The lemma essentially says that transformer can do ``if" clauses (under some assumptions). We make an assumption that $0 \cdot \infty = 0$.

\begin{lemma} \label{lem:if clause for non-neg entries}
    Let $\x \in \R^m$ and $\y, \z \in \R_{\geq 0}^n$, and let $b \in \{0, 1\}$. There exists a two-layer feedforward network $f : \R^{m + 2n +1} \to \R^{m+n}$ such that
    \begin{equation*}
        f([\x, \y, \z, b]) = \begin{cases}
            [\x, \y] &\text{if $b=0$} \\
            [\x, \z] &\text{if $b=1$}.
        \end{cases}
    \end{equation*}    
\end{lemma}

\begin{proof} (Sketch)
    Let $\vec{\infty}$ be the vector of dimension $n$ whose entries are all $\infty$'s.  

    We utilize the 2-layer feedforward network to compute the following:
    $$\ReLU(\y - b \cdot \vec{\infty}) + \ReLU(\z - (1-b) \vec{\infty}),$$
    which equals $\y$ if $b = 0$ and $\z$ if $b = 1$.
\end{proof}

\begin{remark}
    We can implement the feedforward network in the lemma using $2n+1$ additional dimensions (say we use $n$ dimensions to save $\y - b \vec{\infty}$, $n$ dimensions to save $\z - (1-b) \vec{\infty}$, and $1$ to save $1-b$). There can be more optimal usage of dimensions, but we don't need that. 
\end{remark}

Next, we have a lemma that essentially says transformer can multiply a vector $\x$ with a number $a \in \{0, 1\}$. 

\begin{lemma} \label{lem:mult vect by 0 or 1}
    Let $\x \in \R^d$ and $a \in \{0, 1\}$ together in a vector, denoted $\begin{bmatrix}
        \x &a &\0_d &\0_d
    \end{bmatrix}^T$. Then, there exists a 2-layer feedforward network (with ReLU) that returns $\begin{bmatrix}
        \0_d &\_ &(\_)_d &a \x
    \end{bmatrix}$. 
\end{lemma}

\begin{proof}
    First, use the first affine linear transformation to get $$\begin{bmatrix}
        \0_{d} &a &\x - \vec{\infty}(1-a) &-\x - \vec{\infty}(1-a)    
    \end{bmatrix}^T.$$ 
    Next, after ReLU, we get $$\begin{bmatrix}
        \0_d &a &\ReLU(\x -\vec{\infty} (1-a)) &\ReLU(-\x - \vec{\infty} (1-a))
    \end{bmatrix}.$$ Then, we use the second linear transformation to get 
    \begin{equation*}
        \ReLU(\x -\vec{\infty} (1-a)) - \ReLU(-\x - \vec{\infty} (1-a)) = \begin{cases}
            1 \cdot \x &\text{if $a=1$} \\
            0 \cdot \x &\text{if $a=0$}.
        \end{cases}
    \end{equation*}
    Thus, we yield $\begin{bmatrix}
        \0_d &\_ &(\_)_d &a \x
    \end{bmatrix}$, 
    as desired. 
\end{proof}

\begin{remark} We have some remarks on the previous lemma:
    \begin{itemize}
        \item Above, we could have done calculation on the first $d$ dimensions, but we will leave them equal to zero, and do calculations in some $d$ placeholder dimensions. The reason is that when we use feedforward network in a transformer, after all calculations, the transformer performs a skip-connection to add the original vector into the resulting vector. In our case, the first $d$ dimensions will be added by $\x$ in the end, and that may alter the results we want. Thus, we perform calculations in placeholder dimensions, which are unaltered by skip-connections.
        
        \item So, to multiply $\x$ by $a$, we have used $2d$-placeholder dimensions ($d$ of which are used to save the results, too).
    \end{itemize}
\end{remark}

Next, we have a lemma that says if each $\x_i$ has $i$ as positional encoding, then for $\x_i$, we can calculate number $\frac{1}{i}$ (using a transformer). 

\begin{lemma} \label{lem:i gives 1/i easily}
    Let $\x_1, \dots, \x_n$ such that $\x_i = \begin{bmatrix}
        i &0 &0
    \end{bmatrix}$. Then, there exists a transformer with 2 layers that returns $\y_i = \begin{bmatrix}
        i &\_ &\frac{1}{i}
    \end{bmatrix}$. 
\end{lemma}

\begin{proof} 
    Let us just write in terms of $3$ dimensions for conciseness. Let's be understood we may need to use more than $3$ dimensions for the purpose of place-holding. Say we are considering query $\x_i$.
    
    First, by Lemma \ref{lem:tran checks if c_i = t}, we use a layer (layer $1$) to get $\x_i^{(1)} = \begin{bmatrix}
        i &\1(i=1) &0
    \end{bmatrix}$. 
Next, use a head in layer $2$ to attend uniformly over all tokens (with causal mask) with value $\1(j = 1)$. From this, we get $\frac{1}{i}$ as the result, hence obtaining $\x_i^{(2)} = \begin{bmatrix}
        i &\_ &\frac{1}{i}
    \end{bmatrix}$.  
\end{proof}

\begin{remark}
    By Lemma \ref{lem:i gives 1/i easily}, if we have $i$ as a positional encoding for $\x_i$, we can easily obtain $\frac{1}{i}$ for $\x_i$. Thus, WLOG, we can assume $\frac{1}{i}$ as a number given from positional encoding, too.
\end{remark}



\end{document}